\RequirePackage{fix-cm}
\documentclass[smallextended]{svjour3}       
\smartqed  
\usepackage{graphicx}
\usepackage{amsmath,bm}
\usepackage{mathrsfs}
\usepackage{amssymb}
\usepackage{graphicx} 
\usepackage{subfigure}
\usepackage{algorithm}
\usepackage{algorithmic}
\usepackage{times}
\usepackage{etoolbox}

\usepackage[noadjust]{cite}

\usepackage{amsmath,bm}
\usepackage{mathrsfs}
\usepackage{amssymb}
\usepackage{graphicx} 
\usepackage{subfigure}
\usepackage{algorithm}
\usepackage{algorithmic}
\usepackage{times}
\usepackage{etoolbox}

\usepackage[noadjust]{cite}

\usepackage[english]{babel}
\usepackage[autostyle]{csquotes}

\usepackage{subfigure}

\usepackage[normalem]{ulem}
\usepackage[usenames]{color}
\usepackage[table]{xcolor}

\newtheorem{prop}{Proposition}

\usepackage[normalem]{ulem}

\newcommand{\HS}{\mathcal{H}}

\begin{document}

\title{Deep Kernel Recursive Least-Squares Algorithm}


\author{Hossein Mohamadipanah         \and
Mahdi Heydari         \and
        Girish Chowdhary
}


\institute{Hossein Mohamadipanah \at
	Stanford University\\
	Stanford, CA 94305\\
    \email{hmp@stanford.edu}
           \and
           Mahdi Heydari \at
              	Worcester Polytechnic Institute\\
	Worcester, MA 01609\\
	\email{mheydari1985@gmail.com}
\and
           Girish Chowdhary \at
              	University of Illinois at Urbana Champaign\\
	Urbana, IL 61801\\
	\email{girishc@illinois.edu}
}

\date{Received: date / Accepted: date}

\maketitle

\begin{abstract}
We present a new kernel-based algorithm for modeling evenly distributed multidimensional datasets that does not rely on input space sparsification. The presented method reorganizes the typical single-layer kernel-based model into a deep hierarchical structure, such that the weights of a kernel model over each dimension are modeled over its adjacent dimension. We show that modeling weights in the suggested structure leads to significant computational speedup and improved modeling accuracy.
\end{abstract}

\section{Introduction}
\label{Introduction}

Many natural and man-made phenomena are distributed over large spatial and temporal scales. Some examples include weather patterns, agro-ecological evolution, and social-network connectivity patterns. Modeling such phenomena has been the focus of research during the past decades \cite{Tryfona,Vedula}. Of the many techniques studied, Kernel methods \cite{Bishop}, have emerged as a leading tool for data-driven modeling of nonlinear spatiotemporally varying phenomena. Kernel-based estimators with recur.sive least squares (RLS) (or its sparsified version) learning algorithms represent the state-of-the-art in data-driven spatiotemporal modeling \cite{Sliding,NORMA,Fixed,KRLST,QKLMS,TLAF,GARCIAVEGA2019}. However, even with the successes of these algorithms, modeling of large datasets with significant spatial and temporal evolution remains an active challenge. As the size of the dataset increases, the number of kernels that needs to be utilized begins to increase. This consequently leads to a large kernel matrix size and computational inefficiency of the algorithm \cite{Engel}. The major stream of thought in addressing this problem has relied on the sparsification of the kernel dictionary in some fashion. In the time-invariant case, the Kernel Recursive Least-Squares method (KRLS) scales as $\mathcal{O}(m^2)$, where $m$ is the size of the kernel dictionary. Now suppose our dataset is varying with time, and we add time as the second dimension. Then KRLS cost scales as $\mathcal{O}((mm_1)^2)$, where $m_1$ is the number of time-steps. While sparsification can reduce the size of $m$ and $m_1$ to some extent, it is easy to see that as the dimension of the input space increases, the computational cost worsens.

Over the past two decades, many approaches have been presented to overcome computational inefficiency of  naive KRLS. Sliding-Window Kernel Recursive Least Squares (SW-KRLS) method \cite{Sliding} was developed in which only predefined last observed samples are considered. The Naive Online regularized Risk Minimization (NORMA) algorithm \cite{NORMA} was developed based on the idea of stochastic gradient descent within a feature space. NORMA enforces shrinking of weights over samples so that the oldest bases play a less significant role and can be discarded in a moving window approach. Naturally, the main drawback of Sliding-Window based approaches is that they can forget long-term patterns as they discard old observed samples. The alternative approach is to discard data that is least relevant. For example, in \cite{Fixed}, the Fixed-Budget KRLS (FB-KRLS) algorithm is presented, in which the sample that plays the least significant role, the least error upon being omitted, is discarded. A Bayesian approach is presented in \cite{KRLST} that utilizes confidence intervals for handling non-stationary scenarios with a predefined dictionary size. However, the selection of the budget size requires a tradeoff between available computational power and the desired modeling accuracy. Hence, finding the right budget has been the main challenge in fixed budget approaches and also for large scale datasets a loss in modeling accuracy is inevitable. Quantized Kernel Least Mean Square (QKLMS) algorithm is developed based on vector quantization method to quantize and compress the feature space \cite{QKLMS, Han2019, ZHENG2016}. Moreover, Sparsified KRLS (S-KRLS) is presented in \cite{Engel} which adds an input to the dictionary by comparing its approximate linear dependency to the observed inputs, assuming a predefined threshold. In \cite{Hessian}, a recurrent kernel recursive least square algorithm for online learning is presented in which a compact dictionary is chosen by a sparsification method based on the Hessian matrix of the loss function that continuously examines the importance of the new training sample to utilize in dictionary update of the dictionary according to the importance of measurements, using a predefined fixed budget dictionary. Kernel Least Mean p-Power (KLMP) algorithm is proposed in \cite{KLMP} for systems with non-Gaussian impulsive noises. Distributed kernel adaptive filters are presented in \cite{Gao2015, Bouboulis2018} by applying diffusion-based schemes to the Kernel Least Mean Square (KLMS) for distributed learning over networks.
An online nearest-neighbors approach for cluster analysis (unsupervised learning) is presented in \cite{TLAF} based on the kernel adaptive filtering (KAF) framework. An online prediction framework is proposed in \cite{GARCIAVEGA2019} to improve prediction accuracy of KAF.

Majority of the presented methods in the literature are typically based on reducing the number of training inputs by using a moving window, or by enforcing a ``budget'' on the kernel dictionary size. However, moving window approaches lead to forgetting, while budgeted sparsification leads to a loss in modeling accuracy. Consequently, for datasets over large scale spatiotemporal varying phenomena,  neither moving window nor limited budget approaches present an appropriate solution. In this paper, we present a new kernel-based modeling technique for modeling large scale multidimensional datasets that does not rely on input space sparsification. Instead, we take a different perspective and reorganize the typical single-layer kernel-based model in a hierarchical structure over the weights of the kernel model. The presented structure does not affect the convexity of the learning problem and reduces the need for sparsification and also leads to significant computational speedup and improved modeling accuracy. The presented algorithm called 'Deep Kernel Recursive Least Squares (D-KRLS)' herein and is validated on synthetic and real-world multidimensional datasets where it outperforms the state-of-the-art KRLS algorithms.

A number of authors have also explored non-stationary kernel design and local-region based hyperparameter optimization to accommodate spatiotemporal variations \cite{Plagemann2008,Garg2012}. However, the hyperparameter optimization problem in these methods is non-convex and leads to significant computational overhead. As our results show, the presented method can far outperform a variant of the NOn-stationary Space TIme variable Latent Length scale Gaussian Process (NOSTILL-GP) algorithm \cite{Garg2012}. The Kriged Kalman Filter (KKF) \cite{Kriged} models the evolution of weights of a kernel model with a linear model over time. Unlike KKF, our method is not limited to have a linear model over time and can be extended to multidimensional datasets. The main difference between the presented method with the current stream of works in deep learning \cite{Deep_NN_Survey} and deep Gaussian process \cite{Deep_GP} is the fact that our algorithm utilizes a hierarchic approach to model weights instead of having multiple layers of neurons or Gaussian functions. In addition, the presented algorithm is convex and improves the computational cost.

This paper is organized as follows: In section \ref{Deep Kernel Recursive Least-Squares}, the main algorithm is presented in detail. In section \ref{Experimental Results}, three problem domains are exemplified and the result of the method for modeling them are presented and computational cost is compared to the literature. Finally, conclusions and discussion are given in Section \ref{Conclusion}.

%
\section{Deep Kernel Recursive Least-Squares}
\label{Deep Kernel Recursive Least-Squares}
We begin in subsection \ref{Preliminaries} with an overview of the KRLS method.
Then, the details of the proposed algorithm for two and three dimensional problems are presented in subsections \ref{2D_D-KRLS} and \ref{3D_D-KRLS}, respectively. The algorithm is generalized for high dimensional problems in \ref{General Deep Kernel Recursive Least-Squares}. Finally, the computational efficiency is discussed in subsection \ref{Computational}.


\subsection{Preliminaries}
\label{Preliminaries}
\begin{figure}
	\vskip 0.2in
	\begin{center}
		\centerline{\includegraphics[trim=2.4cm 20cm 10.1cm 2.5cm, clip=true, width=\columnwidth]{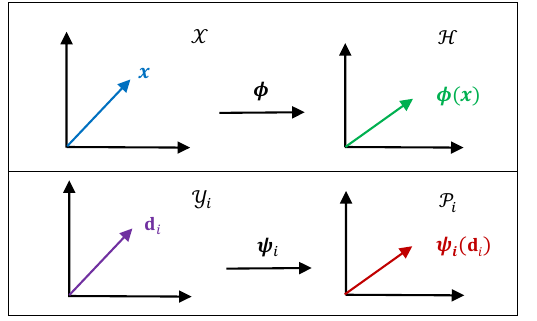}}
		\caption{(Top) mapping inputs $\bm{x}$ to the Hilbert space $\mathcal{H}$ in initial modeling, (Bottom) mapping each hyperparameter vector $\bm{d}_{i}$  to its corresponding Hilbert space $\mathcal{P}_{i}$.}
		\label{Hilbert_HNK}
	\end{center}
	\vskip -0.2in
\end{figure}

Consider a recorded set of input-output pairs $(\bm{z}_1,y_1), (\bm{z}_2,y_2),\dots,(\bm{z}_s,y_s)$, where the input $\bm{z}_{s} \in \mathcal{X}$, for some space $\mathcal{X}$ and $y_{s} \in \mathbb{R}$. By definition, a kernel $ k(\bm{z},\bm{z}^{\prime})$ takes two arguments $\bm{z}$ and $\bm{z}^{\prime}$ and maps them to a real values ($\mathcal{X}\times\mathcal{X} \to \mathbb{R}$). Throughout this paper, $ k(\bm{z},\bm{z}^{\prime})$ is assumed to be continuous. In particular, we focus on Mercer kernels \cite{Mercer}, which are symmetric, positive semi-definite kernels. Therefore for any finite set of points $(\bm{z}_1, \bm{z}_2,\dots,\bm{z}_s)$, the Gram matrix $ [K]_{ij}= k(\bm{z}_i,\bm{z}_j)$ is a symmetric positive semi-definite matrix. Associated with Mercer kernels there is a reproducing kernel Hilbert space $\HS$ and a mapping $ \bm{\phi} :\mathcal{X} \to \HS$ such that kernel $ k(\bm{z},\bm{z}^{\prime})= \langle \bm{\phi}(\bm{z}),\bm{\phi}(\bm{z}^{\prime}) \rangle_\HS $, where $\langle . , . \rangle_\HS$ denotes an inner product in $\HS$, shown in Figure~\ref{Hilbert_HNK} (Top). The corresponding weight vector
$\bm{\omega} = (\omega_{1},\omega_{2},\dots,\omega_{s})^{T}$ can be found by minimizing the following quadratic loss function:
\begin{equation}
\ell(\bm{\omega}) = \sum_{i=1}^{s} (f(\bm{z}_{i})-y_{i})^{2} =  \| K\bm{\omega}-\bm{y} \|^{2}.
\label{eq1}
\end{equation}

The KRLS algorithm has quadratic computational cost  $\mathcal{O}(m^{2})$, where $m$ denotes the number of samples of the input vector \cite{Engel}. Let the number of samples in each dimension be denoted by $m_{k} (k=0,1,\dots,n)$ for an $n+1$ dimensional system, then the algorithm cost is $\mathcal{O}((m_{0} m_{1}\dots m_{n})^{2})$, which can become quickly intractable.

\subsection{2D Deep Kernel Recursive Least-Squares}
\label{2D_D-KRLS}
\begin{figure}[!ht]
	\begin{center}
		\centerline {\includegraphics[trim=2cm 11.5cm 1.5cm 3cm, clip=true, width=\columnwidth]{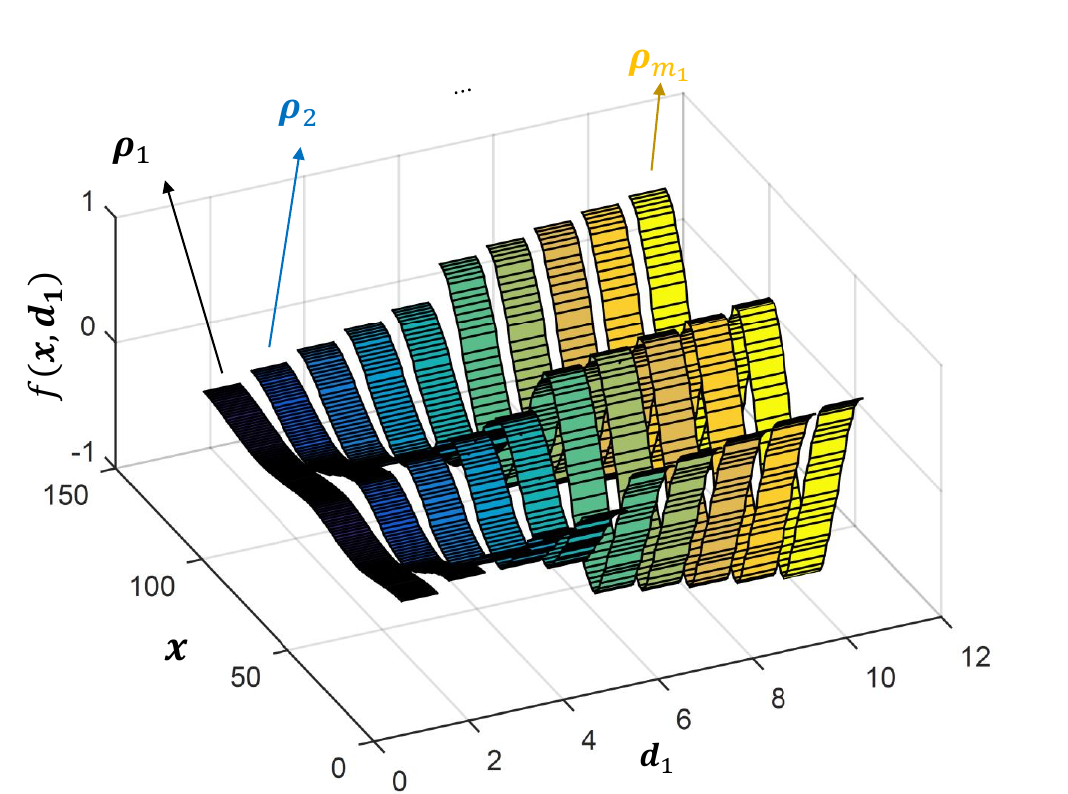}}
		\caption{The two-dimensional function. Training is performed at each $\bm{d}_1$ sample and the corresponding weights ($\bm{\rho}_{1}$ to $\bm{\rho}_{m_{1}}$) are recorded.}
		\label{Ribbon_ieee}
	\end{center}
\end{figure}

Assume that the intention is to model a function with a two-dimensional input, denoted by $\bm{x}$ and $\bm{d}_1$ and one-dimensional output, denoted by $\bm{y}$. The output of this function is a function of the inputs $\bm{x}$ and $\bm{d}_1$, $ f(\bm{x}, \bm{d}_1)$, and the objective of the modeling is to find this function, this is depicted in Figure~\ref{Ribbon_ieee}.

\begin{figure*}[!ht]
	\begin{center}
		\centerline {\includegraphics[trim=2cm 23cm 2cm 3cm, clip=true, width=\columnwidth]{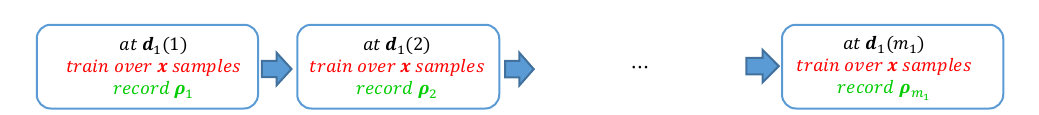}}
		\caption{Training over all samples of $\bm{x}$ at each sample of $\bm{d}_1$ and recording the corresponding weights.}
		\label{2D_Initial}
	\end{center}
\end{figure*}

In the first step, the modeling is performed on all the samples of $\bm{x}$ at the first sample of the $\bm{d}_1$ and the corresponding weight $\bm{\rho}_1$  is recorded. Then modeling is performed on all the samples of $\bm{x} $ at the next sample of $\bm{d}_1$. This process is continued until the last sample of $\bm{d}_1$, illustrated in Figure~\ref{2D_Initial}.

\begin{figure*}[!ht]
	\begin{center}
		\centerline {\includegraphics[trim=2.5cm 14cm 2.5cm 5cm, clip=true, scale = 0.8, width=\columnwidth]{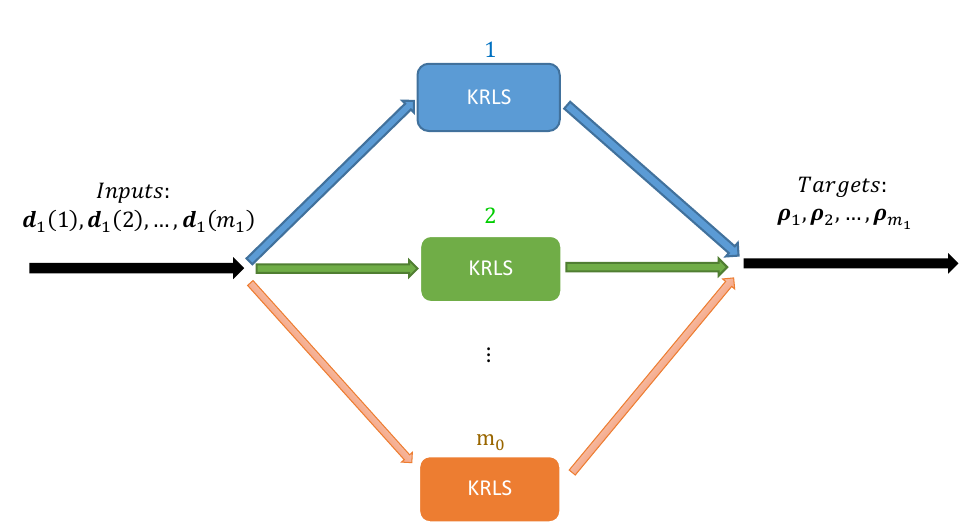}}
		\caption{Modeling $ \bm{\rho}_1 $ to $ \bm{\rho}_{m_1} $ by $m_0$ KRLS since dimension of the output is $m_{0} \times 1$.}
		\label{2D_Alpha}
	\end{center}
\end{figure*}

After recording $\bm{\rho}_1$ to $\bm{\rho}_{m_{1}}$, we put together these vectors and get a matrix $P$ as \eqref{eq2}.
\begin{equation}
P = \begin{bmatrix}
[\bm{\rho}_1]_{m_{0} \times 1} & [\bm{\rho}_2]_{m_{0} \times 1} & \dots & [\bm{\rho}_{m_{1}}]_{m_{0} \times 1}\\
\end{bmatrix}_{m_{0} \times m_{1}}.
\label{eq2}
\end{equation}
The second step is to model each row of the matrix $P$ over $\bm{d}_1$ samples. To perform modeling, $m_{0}$ KRLS models are used as shown in Figure~\ref{2D_Alpha}. Therefore, the dimension of the output is $m_{0} \times 1$. Accordingly, the corresponding weight $\xi$ recorded with dimension $m_{1} \times m_{0}$. This is the end of the training process.

The next step is to validate the model using a validation dataset. At the validation sample $j$ of $\bm{d}_1$ we can estimate $ \Im_{j}$ by \eqref{eq3}.
\begin{equation}
\Im_{j} = \begin{bmatrix}
k_{H1}(\bm{d}_1(1),\bm{d}_{1val}(j)) \\
k_{H1}(\bm{d}_1(2),\bm{d}_{1val}(j)) \\
\vdots\\
k_{H1}(\bm{d}_1(m_{1}),\bm{d}_{1val}(j))\\
\end{bmatrix}_{m_{1} \times 1}
\label{eq3}
\end{equation}
where $ k_{H1} $ is the kernel function which is assumed to be Gaussian with the standard deviation $\sigma_{H1}$. Then, by using \eqref{eq4} and \eqref{eq5}, we can estimate $\bm{\rho}_j$ and $P$, respectively.
\begin{equation}
[ \bm{\hat\rho}_{j}]_{m_{0} \times 1} = [\xi]^{T}_{m_{0} \times m_{1}}[\Im_{j}]_{m_{1} \times 1}.
\label{eq4}
\end{equation}
\begin{equation}
\hat P = \begin{bmatrix}
[\bm{\hat\rho}_1]_{m_{0} \times 1} & [\bm{\hat\rho}_2]_{m_{0} \times 1} & \dots & [\bm{\hat\rho}_{m_{1}}]_{m_{0} \times 1}\\
\end{bmatrix}_{m_{0} \times m_{1val}}.
\label{eq5}
\end{equation}
Then, at the validation sample $s$ of $\bm{x}$, we can calculate $\wp_{s}$ by using \eqref{eq6}.
\begin{equation}
\wp_{s} = \begin{bmatrix}
k_{I}(\bm{x}(1),\bm{x}_{val}(s)) \\
k_{I}(\bm{x}(2),\bm{x}_{val}(s)) \\
\vdots\\
k_{I}(\bm{x}(m_{0}),\bm{x}_{val}(s))\\
\end{bmatrix}_{m_{0} \times 1}
\label{eq6}
\end{equation}
where $ k_{I} $ is the kernel function which is assumed to be Gaussian with the standard deviation $\sigma_{I}$. In consequence, we can estimate the function $f(\bm{x},\bm{d}_1) $ at any validation sample by \eqref{eq7}.
\begin{equation}
f(s,j)= [{\bm{\hat\rho}_{j}}]^{T}_{1 \times m_{0}}[\wp_{s}]_{m_{0} \times 1}.
\label{eq7}
\end{equation}
The experimental results of this algorithm is presented in the subsections \ref{2D} and \ref{Temperature Modeling on Intel Lab Dataset}.

\subsection{3D Deep Kernel Recursive Least-Squares}
\label{3D_D-KRLS}
Consider a function with a three-dimensional input, denoted by $\bm{x}$, $\bm{d}_1$, and $\bm{d}_2$ and one-dimensional output, denoted by $\bm{y}$. The output is a function of the inputs $\bm{x}$, $\bm{d}_1$, and $\bm{d}_2$, and the aim of modeling is to find the function $ f(\bm{x}, \bm{d}_1,  \bm{d}_2)$.

\begin{figure*}[ht]
	\begin{center}
		\centerline {\includegraphics[trim=2cm 12.6cm 2cm 1.8cm, clip=true, width=\columnwidth]{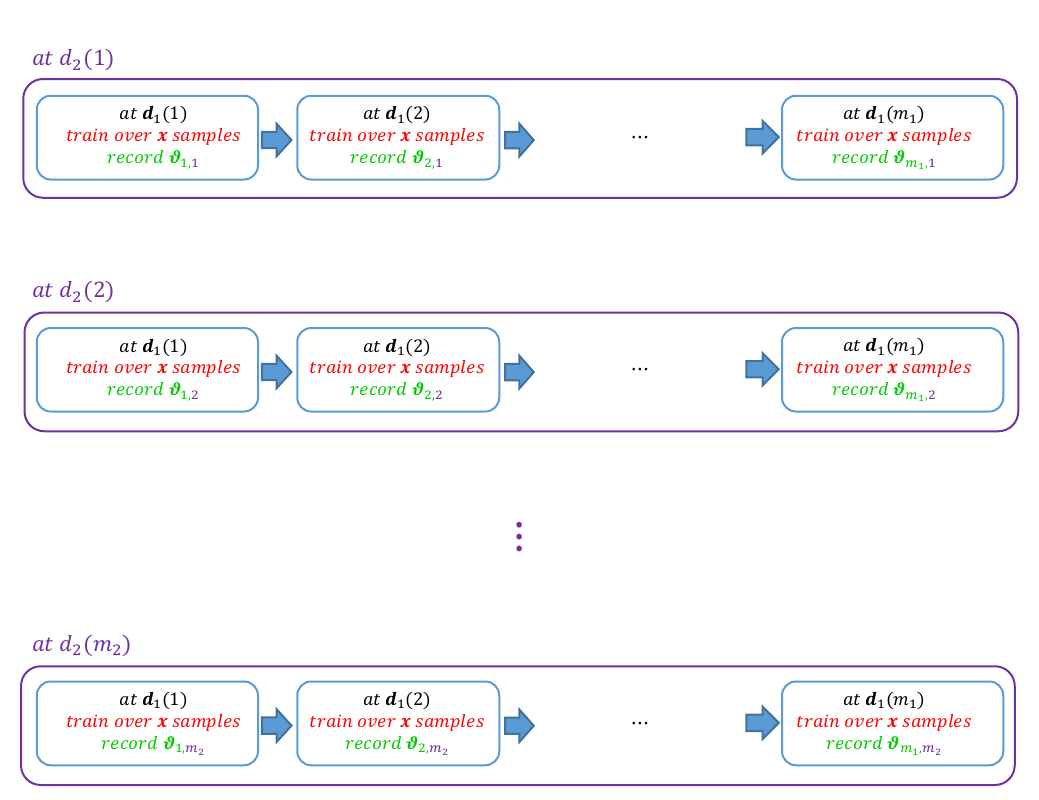}}
		\caption{Training over all samples of $\bm{x}$ at each sample of $\bm{d}_1$ and at each sample of $\bm{d}_2$.}
		\label{3D_Initial}
	\end{center}
\end{figure*}

To execute modeling, at the first sample of $ \bm{d}_1$ the modeling is performed on all samples of $\bm{x}$ and the corresponding weight $\bm{\vartheta}_{1,1}$ is recorded, with dimension $m_0 \times 1$. Then modeling is performed on all samples of $\bm{x} $ at the next sample of $\bm{d}_1$ and the corresponding weight $\bm{\vartheta}_{2,1}$ is recorded. This process is continued until the last sample of $\bm{d}_1$ and recording the weight $\bm{\vartheta}_{m_{1},1}$. Next, this process is repeated for all samples of $\bm{d}_2$, illustrated in Figure~\ref{3D_Initial}, and the corresponding weights are recorded.

\begin{figure*}[!ht]
	\begin{center}
		\centerline {\includegraphics[trim=2.5cm 14cm 1.5cm 5cm, clip=true, scale = 0.8, width=\columnwidth]{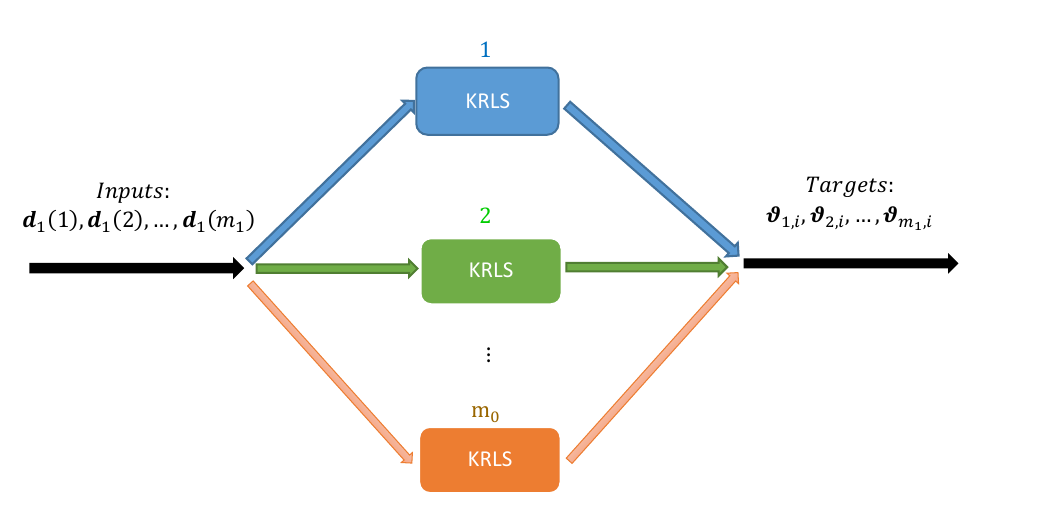}}
		\caption{Modeling column $i$ of $\Theta$ by $m_{0}$ KRLS models.}
		\label{3D_Alpha_Middle}
	\end{center}
\end{figure*}

After recording $\bm{\vartheta}_{1,1}$ to $\bm{\vartheta}_{m_{1},m_{2}}$, we put together these weight vectors and we made a cell $\Theta$ according to \eqref{eq8}. Consider that all of the elements of this cell are the weight vectors with dimension $m_{0} \times 1$.
\begin{equation}
\Theta= \begin{Bmatrix}
[\bm{\vartheta}_{1,1}] & [\bm{\vartheta}_{1,2}] & \dots & [\bm{\vartheta}_{1,m_{2}}]\\
[\bm{\vartheta}_{2,1}] & [\bm{\vartheta}_{2,2}] & \dots & [\bm{\vartheta}_{2,m_{2}}]\\
\vdots & \vdots & \ddots & \vdots\\
[\bm{\vartheta}_{m_{1},1}] & [\bm{\vartheta}_{m_{1},2}] & \dots & [\bm{\vartheta}_{m_{1},m_{2}}]\\
\end{Bmatrix}_{m_{1} \times m_{2}}.
\label{eq8}
\end{equation}
The second step is to model $\Theta$. Each column of $\Theta$ is correlated to one $\bm{d}_2$ sample and the desire is to model each of the columns one by one. Modeling each column of $\Theta$ is illustrated in Figure~\ref{3D_Alpha_Middle}, which is done by $m_{0}$ KRLS since the dimension of each target vector is $ m_{0} \times 1$.

\begin{figure*}[!ht]
	\begin{center}
		\centerline {\includegraphics[trim=2cm 23cm 2cm 1.8cm, clip=true, width=\columnwidth]{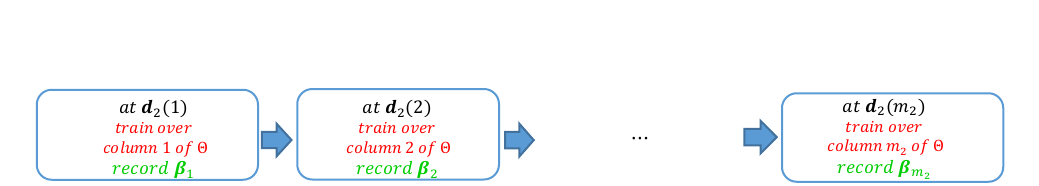}}
		\caption{Training over column $i$ of $\Theta$ at each sample of $\bm{d}_2$ and recording the corresponding weights for $i = 1, 2, \dots, m_{2}$.}
		\label{3D_Middle}
	\end{center}
\end{figure*}

After performing the modeling at each sample of $\bm{d}_{2}$, the corresponding weight $\beta_1$ to $\beta_{m_{2}}$ are recorded, the dimension of each weight is $ m_{1} \times m_{0}$. This process is shown in Figure~\ref{3D_Middle}.

\begin{figure*}[!ht]
	\begin{center}
		\centerline {\includegraphics[trim=2.5cm 14cm 1.5cm 5cm, clip=true, scale = 0.8, width=\columnwidth]{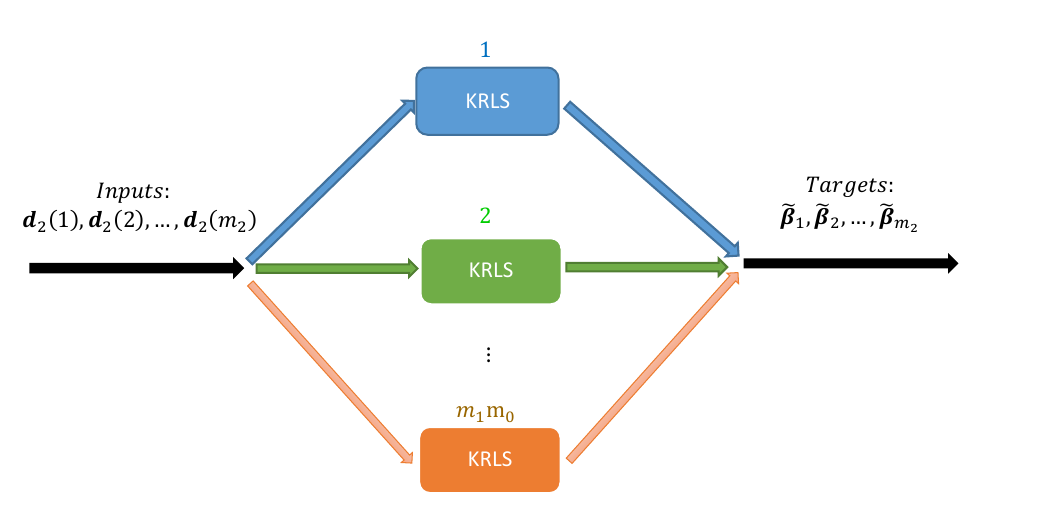}}
		\caption{Modeling $ \bm{\tilde\beta}_1 $ to $ \bm{\tilde\beta}_{m_2} $ by $m_1m_0$ KRLS since dimension of the output is $m_{1}m_{0} \times 1$.}
		\label{3D_Alpha_Last}
	\end{center}
\end{figure*}

After recording $\beta_1$ to $\beta_{m_{2}}$, we put together these vectors and we made a cell as shown in \eqref{eq9}.
\begin{equation}
\Omega = \{[\beta_1]_{m_{1} \times m_{0}}, [\beta_2]_{m_{1} \times m_{0}}, \dots, [\beta_{m_{2}}]_{m_{1} \times m_{0}}\}_{1 \times m_{2}}.
\label{eq9}
\end{equation}
The third step is to model $\Omega$. Each element of $\Omega$ ($\beta_{i} (i = 1,2,\dots, m_2) $) is a $m_{1} \times m_{0}$ matrix. As we need outputs to be in a vector format in KRLS modeling, we defined $\tilde{\beta}_i$ for $i = 1,2,\dots, m_2$ and $ \tilde{\Omega}  $ as presented in \eqref{eq10} and \eqref{eq11} respectively.
\begin{equation}
\tilde{\beta}_i = \begin{bmatrix}
\beta_{i}(:,1) \\
\beta_{i}(:,2) \\
\vdots\\
\beta_{i}(:,m_{0})\\
\end{bmatrix}_{m_{1}m_{0} \times 1}
\label{eq10}
\end{equation}
where $(:,i)$ for $i = 1,2,\dots, m_2$ denotes the column $i$th of a matrix.
\begin{equation}
\tilde{\Omega} = [\tilde{\beta}_1, \tilde{\beta}_2, \dots, \tilde{\beta}_{m_{2}}]_{m_{1}m_{0} \times m_{2}}.
\label{eq11}
\end{equation}
Consider that each column of $\tilde{\Omega}$ is corresponded to one $\bm{d}_2$ sample and the intension is to model them one by one. Modeling of each column of $\tilde{\Omega}$ is illustrated in Figure~\ref{3D_Alpha_Last}.

After performing the modeling, the corresponding weight $\gamma$ is recorded. The dimension of $\gamma$ is $ m_{2}\times m_{1}m_{0}$. At this step modeling is finalized.\\
To validate the model, at the validation sample $i$ of $\bm{d}_2$ we can calculate $\eta_{i}$ by \eqref{eq12}.
\begin{equation}
\eta_{i} = \begin{bmatrix}
k_{H2}(\bm{d}_2(1),\bm{d}_{2val}(i)) \\
k_{H2}(\bm{d}_2(2),\bm{d}_{2val}(i)) \\
\vdots\\
k_{H2}(\bm{d}_2(m_{2}),\bm{d}_{2val}(i))\\
\end{bmatrix}_{m_{2} \times 1},
\label{eq12}
\end{equation}
where $ k_{H2} $ is the kernel function, which is assumed to be Gaussian with the standard deviation $\sigma_{H2}$. Then we estimate the weight $\tilde{\beta}_i $ by using \eqref{eq13}.
\begin{equation}
[\hat{\tilde{\beta}}_i]_{m_{1}m_{0} \times 1} = [\gamma]^{T}_{m_{1}m_{0} \times m_{2}}[\eta_{i}]_{m_{2} \times 1}.
\label{eq13}
\end{equation}
by reshaping $ \hat{\tilde{\beta}}_i $ according to \eqref{eq10}, we can find estimation of $\beta_{i}$, denoted by $\hat{\beta}_{i}$. Thus, we can find estimation of $\Omega$ by \eqref{eq14}.
\begin{equation}
\begin{split}
\hat{\Omega} = \{[\hat{\beta_1}]_{m_{1} \times m_{0}}, [\hat{\beta_2}]_{m_{1} \times m_{0}}, \dots,\\ [\hat{\beta}_{m_{1}}]_{m_{1} \times m_{0}}\}_{1 \times m_{2val}}.
\label{eq14}
\end{split}
\end{equation}
Next, at the validation sample $j$ of $\bm{d}_1$ we can calculate $\tau_{j}$ by \eqref{eq15}.
\begin{equation}
\tau_{j} = \begin{bmatrix}
k_{H1}(\bm{d}_1(1),\bm{d}_{1val}(j)) \\
k_{H1}(\bm{d}_1(2),\bm{d}_{1val}(j)) \\
\vdots\\
k_{H1}(\bm{d}_1(m_{1}),\bm{d}_{1val}(j))\\
\end{bmatrix}_{m_{1} \times 1},
\label{eq15}
\end{equation}
where $ k_{H1} $ is the kernel function, which is assumed to be Gaussian with the standard deviation $\sigma_{H1}$. Then, we estimate the weight $\bm{\vartheta}_{i,j} $ by using \eqref{eq16} and consequently, we can find $\hat\Theta$ by \eqref{eq17}.
\begin{equation}
[ \bm{\hat\vartheta}_{i,j}]_{m_{0} \times 1} = [\hat{\beta_i}]^{T}_{m_{0} \times m_{1}}[\tau_{j}]_{m_{1} \times 1}.
\label{eq16}
\end{equation}
\begin{equation}
\hat\Theta = \begin{Bmatrix}
[\bm{\hat\vartheta}_{1,1}] & [\bm{\hat\vartheta}_{1,2}] & \dots & [\bm{\hat\vartheta}_{1,m_{2}}]\\
[\bm{\hat\vartheta}_{2,1}] & [\bm{\hat\vartheta}_{2,2}] & \dots & [\bm{\hat\vartheta}_{2,m_{2}}]\\
\vdots & \vdots & \ddots & \vdots\\
[\bm{\hat\vartheta}_{m_{1},1}] & [\bm{\hat\vartheta}_{m_{1},2}] & \dots & [\bm{\hat\vartheta}_{m_{1},m_{2}}]\\
\end{Bmatrix},
\label{eq17}
\end{equation}
the dimension of $ \hat\Theta $ is $ m_{1val} \times m_{2val} $. Then, at the validation sample $s$ of $\bm{x}$ we can calculate $\iota_{s}$ by \eqref{eq18}.
\begin{equation}
\iota_{s} = \begin{bmatrix}
k_{I}(\bm{x}(1),\bm{x}_{val}(s)) \\
k_{I}(\bm{x}(2),\bm{x}_{val}(s)) \\
\vdots\\
k_{I}(\bm{x}(m_{0}),\bm{x}_{val}(s))\\
\end{bmatrix}_{m_{0} \times 1},
\label{eq18}
\end{equation}
where $ k_{I} $ is the kernel function, which is assumed to be Gaussian with the standard deviation $\sigma_{I}$. Then we estimate the weight $f(\bm{x},\bm{d}_1,\bm{d}_2) $ at any validation sample by using \eqref{eq19}.
\begin{equation}
f(s,j,i)= [\bm{\hat\vartheta}_{i,j}]^{T}_{1 \times m_{0}}[ \iota_{s}]_{m_{0} \times 1}.
\label{eq19}
\end{equation}
The implementation of this algorithm is in the subsection \ref{3D}.

\subsection{General Deep Kernel Recursive Least-Squares}
\label{General Deep Kernel Recursive Least-Squares}
In this part, we expand the presented idea for modeling higher dimensional datasets. The objective of the modeling is to find the function $ f(\bm{x}, \bm{d}_1,  \bm{d}_2, \dots, \bm{d}_n)$ in which inputs, denoted by $\bm{x}$, $\bm{d}_1$, $\bm{d}_2$ to $\bm{d}_n$.

\begin{figure*}[!ht]
	\begin{center}
		\centerline {\includegraphics[trim=0cm 21.1cm 0cm 1.5cm, clip=true, scale = 0.9, width=\columnwidth]{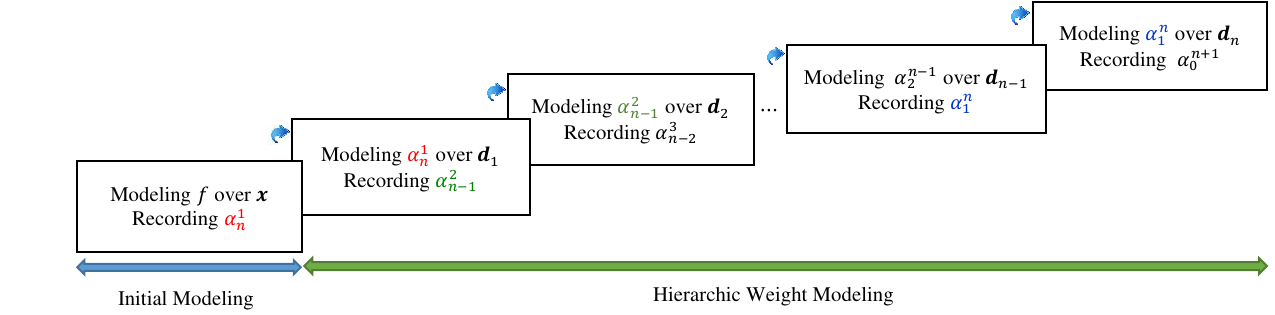}}
		\caption{The structure of the D-KRLS method.}
		\label{Upper_Hyper}
	\end{center}
\end{figure*}

Figure~\ref{Upper_Hyper} illustrates the structure of the D-KRLS method in a general form. Consider a function with $n+1$ dimensional input, represented by $ \bm{x}, \bm{d}_{1}, \bm{d}_{2},\dots,\bm{d}_{n}$, and with $m_{k} (k = 0, 1, \dots, n)$ samples in each dimension. The key idea is to perform training in multiple hierarchical steps. The algorithm consists of two major steps. The first step is called \textit{Initial Modeling} in which modeling is performed over the first dimension ($\bm{x}$) and the corresponding weights are recorded at any sample of $\bm{d}_{1}$ to $\bm{d}_{n}$.

\begin{figure*}
	\centering
	\mbox{\subfigure{\includegraphics[trim=2cm 19cm 31cm 2cm, clip=true, scale = 0.6]{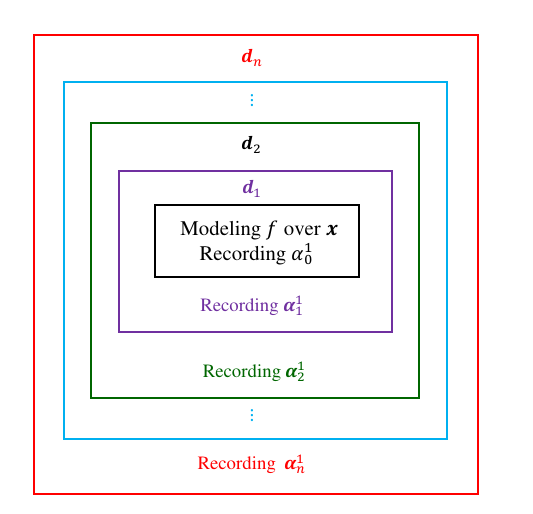}}\quad
		\subfigure{\includegraphics[trim=2cm 19cm 31cm 2cm, clip=true, scale = 0.6]{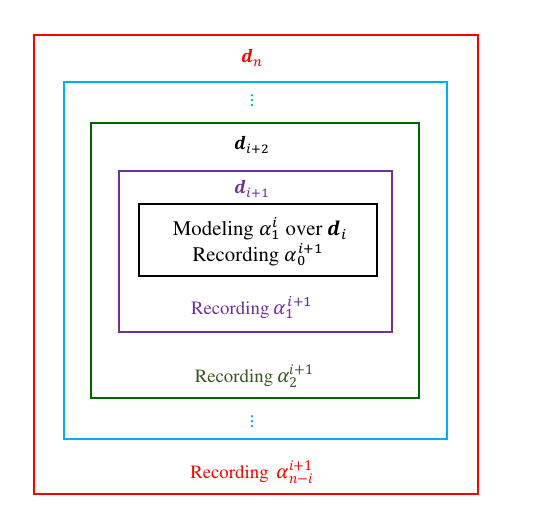}}}
	\caption{(Left) Initial Modeling; (Right) Hierarchic Weight Modeling.}
	\label{Both_Intial_Weight}
\end{figure*}

First $ f$ is trained by KRLS and the corresponding weight vector $\bm{\alpha}_0^1$ is found, Figure~\ref{Both_Intial_Weight} (Left). This training is performed $ m_{0} \times m_{1} \times \dots \times m_{n} $ times to achieve complete model of $f$ at any sample of $\bm{d}_{1}$ to $\bm{d}_{n}$ by solving the optimization in \eqref{eq20} where $ \bm{K}^{1} $ is the Gram matrix associated to the dimension $\bm{x}$ and where $ \bm{y}_{k} $ is the desired target vector for $k = 1, 2, \dots, n+1$. Corresponding to Algorithm~\ref{Algo} ($i = 0$), the superscript of the weight ($\alpha$) represents step and the subscript is the number of associated dimension that the corresponding weight is the result of training on those dimensions (i.e. number $0$ is used for the cases when the weight is not trained over any dimension).
\begin{equation}
\ell'({\alpha}_{k-1}^{1}) = \min_{{\alpha}_{k-1}^{1}}\| {K}^{1}{\alpha}_{k-1}^{1}-\bm{y}_{k} \|^{2},
\label{eq20}
\end{equation}
The cost of Initial Modeling is $\mathcal{O}(m_{n} m_{n-1}\dots m_{1} (m_{0})^{2})$ .

\begin{algorithm}[tb]
	\caption{D-KRLS Algorithm}
	\label{Hyper}
	\begin{algorithmic}
		\FOR{$i=0$ {\bfseries to} $n$}
		\FOR{$k=1$ {\bfseries to} $n +1 - i$}	
		\IF{$i = 0$} \STATE {\bfseries Initial Modeling}  	
		\STATE $ \min_{{\alpha}_{k-1}^{1}}\| {K}^{1}{\alpha}_{k-1}^{1}-\bm{y}_{k} \|^{2}$		
		\ELSIF{$i>0$} \STATE {\bfseries Hierarchic Weight Modeling}	
		\STATE $\min_{{\alpha}_{k-1}^{i+1}}\| {K}^{i+1}{\alpha}_{k-1}^{i+1}-{\alpha}_{k}^{i} \|^{2} $
		\ENDIF
		\ENDFOR
		\STATE {\bfseries Record:} $\alpha_{n-i}^{i + 1}$
		\ENDFOR
	\end{algorithmic}
	\label{Algo}
\end{algorithm}

The second step is called \textit{Hierarchic Weight Modeling} as shown in Figure~\ref{Both_Intial_Weight} (Right). In this step, first the weight $\bm{\alpha}_{1}^i$, is modeled over $\bm{d}_{i}$ for $ i = 1, 2, \dots, n$. Assume $\bm{d}_{i} \in \mathcal{Y}_{i}$, for some space $\mathcal{Y}_{i}$. Therefore, there exist another Hilbert Space $\mathcal{P}_{i}$, and a mapping $ \bm{\psi}_{i} : \mathcal{Y}_{i} \to \mathcal{P}_{i}$ such that $ k(\bm{d}_{i},\bm{d}_{i}^{\prime}) = \langle \bm{\psi}_{i}(\bm{d}_{i}),\bm{\psi}_{i}(\bm{d}_{i}^{\prime}) \rangle_{\mathcal{P}_{i}} $, as shown in Figure~\ref{Hilbert_HNK} (Bottom). The training is performed by minimizing the loss function \eqref{eq21}:
\begin{equation}
\ell''({\alpha}_{k-1}^{i+1}) = \min_{{\alpha}_{k-1}^{i+1}}\| {K}^{i+1}{\alpha}_{k-1}^{i+1}-{\alpha}_{k}^{i} \|^{2},
\label{eq21}
\end{equation}
where $k = 1, 2, \dots, (n+1-i)$ and $i = 1, 2, \dots, n$. It should be mentioned that the desired learning target in this step is the recorded weight from the previous step, namely $ \bm{\alpha}_{k}^{i}$, which is correspond to the Algorithm~\ref{Algo} ($i>0$).

In summary, to model $ f(\bm{x}, \bm{d}_1,  \bm{d}_2, \dots, \bm{d}_n)$ it is required to train the corresponding weight $\alpha^{1}_{n}$. To model $\alpha^{1}_{n}$, it is required to train the corresponding weight $\alpha^{2}_{n-1}$ and to model $\alpha^{2}_{n-1}$ it is required to train the corresponding weight $\alpha^{3}_{n-2}$. This process continues until $\alpha^{n}_{1}$ is modeled, illustrated in Figure~\ref{Upper_Hyper}. The presented method uses a recursive formula and it has to be batch over the first $n$ dimensions, for a $n+1$ dimensional dataset.

\subsection{Computational Efficiency of the D-KRLS Method}
\label{Computational}
Although the approach in the above subsection seems complicated at first glance, it is in fact significantly more efficient compared to the KRLS method. The main reason is that the D-KRLS algorithm divides the training procedure into multiple steps, shown in Figure~\ref{Upper_Hyper}, and utilizes smaller sized kernel matrices instead of using one large sized kernel matrix. The total computational cost of the D-KRLS algorithm for a $n+1$ dimensional dataset is $ \mathcal{O}(m_{n}m_{n-1}\dots m_{1}(m_{0})^2 + m_{n}m_{n-1}\dots m_{2}(m_{1})^2 + m_{n}m_{n-1}\dots m_{3}(m_{2})^2 + \dots + m_{n}(m_{n-1})^2 + (m_{n})^2)$, which is significantly less than the KRLS method cost, $\mathcal{O}((m_{0} m_{1}\dots m_{n})^2)$:\\
\begin{prop}
	\label{prop}
	\textit{Let $\bar{A} = m_{n}m_{n-1} \dots m_{1}(m_{0})^2 + m_{n}m_{n-1} \dots m_{2}(m_{1})^2 + \dots + m_{n}(m_{n-1})^2 + (m_{n})^2$ denotes the computational cost of D-KRLS and $\bar{B} = (m_{0} m_{1}\dots m_{n})^2$ denotes the cost for KRLS. If the number of samples $m_{i}\geq2, i\in(1,2,\dots,n)$ and $m_{0} = m_{1} = \dots = m_{n}$, then $\bar{A} < \bar{B}$}.\\
	The proof is presented in Appendix \ref{Proof}.
\end{prop}
Although Proposition \ref{prop} is restricted to $m_{0} = m_{1} = \dots = m_{n}$, the limit of the D-KRLS cost ($\bar{A}$) as $m_{i}$ (for $i=1,2,\dots,n$) approaches infinity, is less than the KRLS cost:
\begin{equation}
\begin{split}
\lim_{m_{i(i= 0, \dots, n)} \to \infty} \mathcal{O}(\bar{A}) = \mathcal{O}(m_{n}m_{n-1}\dots m_{1}(m_{0})^2)\\<<
\mathcal{O}((m_{0} m_{1}\dots m_{n})^2).
\end{split}
\end{equation}

\section{Numerical Experiments}
\label{Experimental Results}
The D-KRLS method is exemplified on two synthetic and a real world datasets in subsections \ref{2D} to \ref{Temperature Modeling on Intel Lab Dataset}. Then, a discussion is presented regarding cross-correlation between space and time in subsection \ref{Space-time cross-correlation}. Finally, in subsection \ref{Comparison and Discussion} performance of the D-KRLS method is compared to the literature. It should be noted that all algorithms were implemented on an $Intel(R) Core(TM)i7-4700MQ CPU @ 2.40GHz$ with $8GB$ of $RAM$ and the Gaussian kernel, $ k(\bm{z},\bm{z}^{\prime}) = exp(-(\bm{z}-\bm{z}^{\prime})^{T}(\bm{z}-\bm{z}^{\prime})/(2\sigma^2))$ is used in running all the algorithms herein.

\subsection{Synthetic 2D Data Modeling}
\label{2D}
Exemplification of the D-KRLS method on a synthetic two-dimensional spatiotemporal function is presented in this subsection. The two-dimensional nonlinear function $ \varUpsilon(\bm{x},\bm{d}) $ is given by:
\begin{equation}
\varUpsilon(\bm{x},\bm{d}) = \sin(\bm{x})cos(\frac{\bm{d}}{2}),
\label{eq_2D}
\end{equation}
where $\bm{x}$ and $\bm{d}$ are arranged to be $145$ and $150$ evenly divided numbers ranging between $[0.1, 4\pi]$ and $[0.1, 8\pi]$ respectively, while the trigonometric functions are in radians. Consequently, by having $145$ and $150$ data points in each of the two directions, the total number of data points is $21750$. To train and validate, this dataset is divided randomly with $80\%$ of the data used for training and $20\%$ for validation in each dimension. For training, there are $116$ points in the $\bm{x}$ direction and $120$ points in the $\bm{d}$ direction and in total $13,920$ data points.

\begin{figure}[ht]
	\vskip 0.2in
	\begin{center}
		\centerline{\includegraphics[trim=2.8cm 10.5cm 2.8cm 6.0cm, clip=true,width=\columnwidth]{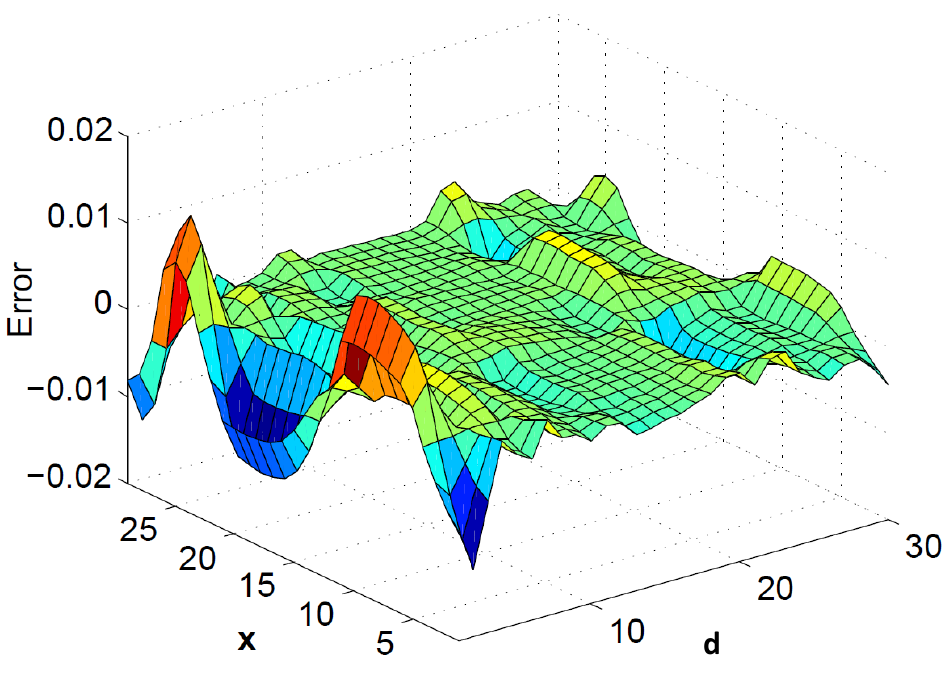}}
		\caption{ Validation error in modeling of the 2D synthetic dataset.}
		\label{2DHNK}
	\end{center}
	\vskip -0.2in
\end{figure}

In the first step of the method, Initial Modeling is performed, using Algorithm 1 ($i = 0$). The kernel matrix size equals $116\times116$ as there are $116$ points in $\bm{x}$ direction and no sparsification is used to demonstrate maximum D-KRLS cost. The variance of the Gaussian kernel is found empirically to be $\sigma_{I} = 1$. The corresponding weight $\bm{\alpha}_0^1$ of this matrix is a vector $116\times1$. Initial Modeling took $2.80$ seconds for all $120$ data samples of $ \bm{d}$. Then modeling of $\bm{\alpha}_1^1$ is done by Algorithm 1 ($i = 1$). The kernel matrix size is equal to $120\times120$ as there are $120$ points in $\bm{d}$ direction. The variance of the Gaussian kernel is found empirically to be $\sigma_{H1} = 0.3$. The Weight Modeling took $0.13$ seconds. The cost of Initial Modeling is more than Hierarchic Weight Modeling, as the Initial Modeling training is done on all samples of $\bm{x}$ at all samples of $\bm{d}$. The total computational time for both steps to train D-KRLS is $2.93$ seconds. The corresponding error for validation is presented in Figure~\ref{2DHNK}. For some data samples of $ \bm{d} $ the error is bigger than the rest, as the same variance of the Gaussian kernel in Initial Modeling ($\sigma_{I}$) is used for all data samples over $ \bm{d} $. Nevertheless, the maximum magnitude of recorded error is found to be $0.0134$, which indicates high validation accuracy.

\subsection{Synthetic 3D Data Modeling}
\label{3D}
To demonstrate the capability of the D-KRLS algorithm in modeling higher dimensional datasets, the presented algorithm is implemented on a synthetic three-dimensional function in this subsection. A three-dimensional function $\Xi(\bm{x},\bm{d}_{1},\bm{d}_{2})$ is defined as follows:
\begin{equation}
\Xi(\bm{x},\bm{d}_{1},\bm{d}_{2}) = \cos(\bm{x})sin(\frac{\bm{d}_{1}}{2})sin(\frac{\bm{d}_{2}}{3}),
\label{eq_3D}
\end{equation}
in which $\bm{x}$, $\bm{d}_1$, and $\bm{d}_2$ are arranged to be $145$, $150$, and $100$ evenly distributed numbers ranging between $[0.1, 4\pi]$, $[0.1, 8\pi]$, and $[0.1, 12\pi]$ respectively, while the trigonometric functions are in radians. Consequently, the total number of the data points is equal to $2,175,000$, ($145\times150\times100$). To train and validate, this dataset is divided randomly with $80\%$ for training and $20\%$ for validation over each dimension. Therefore, there are $1,113,600$ data points for training ($116\times120\times80$) and $17,400$ points ($29\times30\times20$) for validation.

In the first step, Initial Modeling is performed, using Algorithm~\ref{Algo} ($i = 0$). The kernel matrix is equal to $116\times116$ as there are $116$ points in $\bm{x}$ direction. The variance of the Gaussian kernel in Initial Modeling is found empirically to be $\sigma_{I} = 1$. The corresponding weight $\bm{\alpha}_0^1$ is a vector $116\times1$. Initial Modeling took $900.8598$ seconds as the system is trained for all the samples of $\bm{d}_1$ and $\bm{d}_2$. Then modeling of $\bm{\alpha}_2^1$ is done by Algorithm~\ref{Algo} ($i = 1$). The corresponding kernel matrix is equal to $120\times120$ as there are $120$ points in $\bm{d}_1$ direction. It should be noted that the training of $\bm{\alpha}_1^1$ is a vector valued kernel model, as $\bm{\alpha}_0^1$ is a vector $116\times1$ . Consequently, its corresponding weight $\bm{\alpha}_0^2$ is a matrix $120\times116$ and it is formed here as a $13920\times1$ vector. The variance of the Gaussian kernel is found empirically to be $\sigma_{H1} = 0.3$. This process took $8.6569$ seconds as the system is trained for all the samples of $\bm{d}_2$. Then to have the model of $\bm{\alpha}_2^1$, it is required to model $\bm{\alpha}_1^2$ using Algorithm~\ref{Algo} ($i = 2$). The corresponding kernel matrix is equal to $80\times 80$ as there are $80$ points in $\bm{d}_2$ direction and, using $\sigma_{H2} = 1$ as the variance of the Gaussian kernel. This process took $0.7193$ seconds. The cost of training decreases for each step since there is one dimension less in each step compared to its previous step. The total computational time to train the D-KRLS method is $910.2360$ seconds. The corresponding maximum magnitude of the recorded error is found to be $0.0171$, which demonstrates high capability of D-KRLS in modeling of this dataset.

\subsection{Temperature Modeling on Intel Lab Dataset}
\label{Temperature Modeling on Intel Lab Dataset}
In this subsection, the presented algorithm is exemplified on a realistic two-dimensional spatiotemporal environment \cite{Intel} in which $54$ sensors were arranged in Intel lab and the temperature is recorded. It is assumed herein that the sensor indices represent the location of the sensors over space, $\bm{x}$, and time, $\bm{d}$. The time, $\bm{d}$, is arranged from $301$ seconds to $400$ seconds (with $1$ second interval) and $52$ sensors are used (the sensors number $5$ and $15$ are not used to reduce outliers). Consequently, by having $100$ and $52$ data points in each direction, the total number of the data points equals $5200$. To reduce the outliers, the dataset is filtered by a 2D Gaussian filter with variance $5$ and size $6\times6$. To train and validate, this dataset is divided over time randomly with $80\%$ for training and $20\%$ for validation. Totally, there are $4160$ data points for training, and there are $1040$ points for validation.

Therefore the total computational time to train the D-KRLS is $1.11$ seconds. The maximum magnitude of recorded error is found to be $3.11$, as the maximum values of the dataset is $25^\circ$C, the normalized error is $0.1244$.

\subsection{Space-Time Cross-Correlation}
\label{Space-time cross-correlation}
This subsection emphasize on the importance of cross-correlations between dimensions. In the presented algorithm, all of the possible correlations between data points from different dimensions (i.g. space and time) are taken into account and no assumed predefined function is used in modeling of these correlations. In contrast, cross-correlations between space and time have been modeled in the literature by providing different space-time covariance functions \cite{crossfunction}, such as:
\begin{equation}
\mathcal{C}(u,h) = \frac{1}{(a'^{2}u^{2}+1)^{(\frac{1}{2})}} exp(-\frac{b'^{2}h^{2}}{a'^{2}u^{2}+1}),
\label{eqcorrelation}
\end{equation}
in which $a'$ and $b'$ are scaling parameters of time and space respectively. We considered $ \mathcal{C} $ as the input to KRLS, called the NONSTILL-KRLS method herein, inspired from NONSTILL-GP \cite{Garg2012}. In general, appropriate predefined function should result the same level of validation error compared to the KRLS method.

The results in this subsection are found by running the algorithms with $a' =  b' = 1$ on the dataset which is assumed to be a two-dimensional function, denoted in \eqref{eq_2D}, where $\bm{x}$ and $\bm{d}$ are arranged to be $48$ and $50$ evenly distributed numbers ranging between $[0.1, 0.4244]$ and $[0.1, 0.8488]$ respectively, while the trigonometric functions are in radians. To train and validate, this dataset is divided randomly with $80\%$ of the data used for training and $20\%$ for validation in each dimension. For training, there are $37$ points in the $\bm{x}$ direction and $38$ points in the $\bm{d}$ direction and in total $1406$ data points. The reason for using a smaller sized dataset to compare to 2D synthetic dataset, described in subsection \ref{2D}, is that KRLS and NONSTILL-KRLS add all data samples to their dictionary and consequently become extremely expensive and unable to model this large dataset. Hence, we used a smaller dataset in this subsection.

As tabulated in TABLE~\ref{Cross}, NONSTILL-KRLS does not provide an appropriate level of accuracy because of its constraint to define a function in advance. Finding appropriate covariance functions and parameters of such functions is the main challenge. Also, computational time of NONSTILL-KRLS is very high and close to KRLS. On the other hand, the D-KRLS method (similar to KRLS) does not have any predefined model for correlations between space and time and, hence can achieve high level of accuracy.

\begin{table*}[ht]
	\caption{Comparison between D-KRLS, KRLS and NONSTILL-KRLS, in term of computational time in training and corresponding average and maximum validation errors.}
	\vskip 0.15in
	\begin{center}
		\begin{small}
			\begin{sc}
				\begin{tabular}{lccc}
					\hline\\
					\multicolumn{1}{c}{\bf Methods}  &\multicolumn{1}{c}{\bf Training Time}  &\multicolumn{1}{c}{\bf Average Validation Error } &\multicolumn{1}{c}{\bf Maximum Validation Error}\\
					\hline\\
					D-KRLS          & 0.3940 $s$ & 0.0029 & 0.0519 \\
					KRLS                  & 22.6461 $s$ & 0.0035 & 0.0418\\
					NONSTILL-KRLS          &  22.6675 $s$ & 0.0728 & 1.5323 \\
					\hline
				\end{tabular}
			\end{sc}
		\end{small}
	\end{center}
	\label{Cross}
	\vskip -0.1in
\end{table*}

\subsection{Summary of Comparison with Existing Methods}
\label{Comparison and Discussion}
In this section, the D-KRLS is compared with leading kernel-based modeling methods in the literature. Table~\ref{Comparison} presents the comparison in terms of computational training time and maximum validation error of the presented algorithm with the studied kernel adaptive filtering algorithms in \cite{comparative}: QKLMS \cite{QKLMS}, FB-KRLS \cite{Fixed}, S-KRLS \cite{Engel}, SW-KRLS \cite{Sliding}, and NORMA \cite{NORMA} (the codes used here can be found in \cite{vanvaerenberghthesis}).

As detailed in TABLE~\ref{Comparison}, the D-KRLS method resulted in less computational time and also achieved lower maximum validation error compared to the other methods in the literature. For example, it is about five times faster and more accurate than S-KRLS in modeling the Intel Lab dataset. The hyperparameters that are used in running the algorithms are tabulated in TABLE~\ref{Coe} in Appendix \ref{Coefficients}. Since the D-KRLS method results in an improvement in both computational time and accuracy, choosing different hyperparameter sets for the other algorithms, as a trade-off between accuracy and efficiency, does not change the general conclusion of this comparison. The results demonstrates that the D-KRLS method is much more efficient than the other methods, particularly for higher dimensional datasets. For the three-dimensional dataset, the majority of the methods fail and cannot provide a reasonably small validation error. However, D-KRLS models this dataset with high accuracy and less cost compared to all the other methods. Although, it is possible to perform modeling for higher dimensional datasets using D-KRLS, the comparison in TABLE~\ref{Comparison} is limited to three-dimensional as the other methods in the literature fails in modeling of datasets above three-dimensional.

\begin{table*}[ht]
	\caption{Comparison between D-KRLS and other methods in the literature, in terms of computational time in training the datasets and corresponding maximum validation errors.}
	\vskip 0.15in
	\begin{center}
		\begin{small}
			\begin{sc}
				\begin{tabular}{lcccr}
					\hline\\
					\multicolumn{1}{c}{\bf Methods}  &\multicolumn{1}{c}{\bf 2D dataset} &\multicolumn{1}{c}{\bf 3D dataset} &\multicolumn{1}{c}{\bf Intel Lab dataset} \\
					&\multicolumn{1}{c}{13,920 Samples}&\multicolumn{1}{c}{1,113,600 Samples}&\multicolumn{1}{c}{4,160 Samples}\\
					&  (Time | Error) &  (Time | Error) &  (Time | Error) \\
					\hline\\
					\bf D-KRLS          & \bf {2.93 $s$ | 0.0134}     & \bf{910.23 $s$ | 0.0171}      & \bf{1.11 $s$ | 0.1244  }   \\
					QKLMS           & 39.39 $s$ | 0.8401    & 68168.58 $s$ | 0.4469    & 1.54 $s$ | 0.6220    \\
					FB-KRLS         & 285.76 $s$ | 0.0572   & 12994.17  $s$ | 0.9232   & 39.88 $s$ | 0.2182    \\
					S-KRLS          & 108.66 $s$ | 0.0232   & 4608.40 $s$ | 0.8067     & 5.64 $s$ | 0.1582     \\
					NORMA           & 38.61 $s$ | 0.3502    & 2309.76 $s$ | 0.9949     & 1.49 $s$ | 0.3422   \\
					SW-KRLS         & 1097.26 $s$ | 0.9971  & 7696.30 $s$ | 0.9949     & 288.77 $s$ | 0.9895  \\
					\hline
				\end{tabular}
			\end{sc}
		\end{small}
	\end{center}
	\label{Comparison}
	\vskip -0.1in
\end{table*}


\section{Conclusion}
\label{Conclusion}
We presented a kernel method for modeling evenly distributed multidimensional datasets. The proposed approach utilizes a new hierarchic fashion to model weights of each dimension over its adjacent dimension. The presented deep kernel algorithm was compared against a number of leading kernel least squares algorithms and was shown to outperform in both accuracy and computational cost. The method provides a different perspective that can lead to new techniques for scaling up kernel-based models for multidimensional datasets.




%
%



\bibliographystyle{spphys}

\bibliography{ref}

\appendix       

\section{Proof}
\label{Proof}

\begin{proof}
	{The proof is done by induction.

		Let $n = 1$:\\
		$m_{1}(m_{0})^2 < (m_{0}m_{1})^2$ as $m_{0} = m_{1} = ... = m_{n} = m$ and  $m_{i}\geq2$ for $i=1,2,...,n$.  Therefore, $m^3 < m^4$ and the proposition holds for $n = 1$.\\
		Assume for $n = k$ the Proposition holds:\\
		$\bar{A} - \bar{B} < 0$.\\
		Let $n = k + 1$: Therefore the statement $\bar{A} - \bar{B} < 0$ can be written as:

		$m_{k+1}m_{k}m_{k-1}...m_{1}(m_{0})^2)+m_{k+1}m_{k}m_{k-1}...m_{2}(m_{1})^2+...+m_{k+1}(m_{k})^2+(m_{k+1})^2 - (m_{0} m_{1}... m_{k} m_{k+1})^2 < 0$,\\
		$\Rightarrow m_{k+1}\bar{A} + (m_{k+1})^2 - \bar{B}(m_{k+1})^2 < 0$,\\
		as $m_{0} = m_{1} = ... = m_{n} = m$, therefore $m\bar{A} + m^2 - \bar{B}m^2 < 0$,\\
		$\Rightarrow \bar{A}m^2 - \bar{A}m^2 + m\bar{A} + m^2 - \bar{B}m^2 < 0$,\\
		$\Rightarrow m^2(\bar{A}-\bar{B})-\bar{A}m^2 + m\bar{A} + m^2 < 0$,\\
		as $\bar{A} - \bar{B} < 0$, the first term is always negative and therefore it is sufficient to show that summation of the other terms is also negative.\\
		$\Rightarrow -\bar{A}m^2 + m\bar{A} + m^2 < 0$,\\
		$ \Rightarrow m(\bar{A} - m\bar{A}) + m^2 < 0$,\\
		$ \Rightarrow m\bar{A}(1 - m) + m^2 < 0$,\\
		$\hat{A} = m\bar{A}$, $\Rightarrow \hat{A}(1 - m) + m^2 < 0$,\\
		$\Rightarrow \hat{A}(1 - m) < -m^2$,\\
		$\Rightarrow \frac{\hat{A}(-1 + m)}{m^2} > 1$,\\
		$ \Rightarrow (m^{k + 1} + m^{k} + ... + m^{2} + m^{1})(-1 + m) > 1$,\\
		$ \Rightarrow m^{k + 2} - m > 1$,\\
		$ \Rightarrow m(m^{k + 1} - 1) > 1$,\\
		$ \Rightarrow (m^{k + 1} - 1) > \frac{1}{m}$,\\
		As $ m \geq 2$, the right hand side is always equal or less than 0.5 and  the left hand side is always equal or greater than 1, therefore the  statement holds  for $ n = k + 1 $.}	
\end{proof}

\section{Hyperparameters}
\label{Coefficients}
The hyperparameters used in running the algorithms are tabulated in Table~\ref{Coe}.

\begin{table*}[ht]
	\caption{The hyperparameters used in the section III.}
	\vskip 0.15in
	\begin{center}
		\begin{small}
			\begin{sc}
				\begin{tabular}{lcccr}
					\hline\\
					Methods & Synthetic 2D & Synthetic 3D & Intel Lab \\
					\hline\\
					\bf{D-KRLS} &   &   &   \\
					($\sigma_{I}, \sigma_{H1}, (\sigma_{H2})$)  & 1, 0.3 & 1, 0.3, (1) & 1, 1.5\\
					\bf{QKLMS} & & & \\
					($\sigma$, $\mu$) & 1, 0.15& 0.5, 0.03& 3.5, 0.15\\
					($\epsilon$ ) & $10^{-6}$ & 0.0005 & $10^{-6}$ \\
					\bf{FB-KRLS} & & &  \\
					($\sigma$, $D$) & 1, 800 & 0.5, 600 & 3.5, 500 \\
					($\lambda$, $\mu$) & 0.1, 0 & 0.01, 0.03 &  0.01, 0.03 \\
					\bf{S-KRLS} & & & \\
					($\sigma, \delta$) &1, 0.01 & 1, 0.99 & 3.5, 0.2 \\
					\bf{NORMA} & & &  \\
					($\sigma$, $D$) &  1, 13920 & 1, $10^{+4}$ & 3.5, 4160 \\
					($\eta$, $\lambda$) & 0.02, +$10^{-4}$ & 0.005, $10^{-7}$ & 0.04, $10^{-6}$ \\
					\bf{SW-KRLS} & & & \\
					($\sigma$, $D$)  & 1, 1000 & 1, 300 & 3.5, 1000 \\
					($c$)  & 0.01 & 0.01 & 0.01 \\
					\hline
				\end{tabular}
			\end{sc}
		\end{small}
	\end{center}
	\label{Coe}
	\vskip -0.1in
\end{table*}

\end{document}